\documentclass[letterpaper]{article} 
\usepackage[preprint]{aaai2027}  
\usepackage[hyphens]{url}  
\usepackage{graphicx} 
\usepackage{natbib}  
\usepackage{caption} 
\usepackage{booktabs}
\usepackage{amsmath}
\usepackage{xcolor}
\usepackage{tabularx}
\usepackage{array}
\usepackage{placeins}
\usepackage{algorithm}
\usepackage{algorithmic}

\newcolumntype{Y}{>{\raggedright\arraybackslash}X}
\newcolumntype{C}{>{\centering\arraybackslash}X}

\newenvironment{appcoltable}
  {\begin{table}[!htbp]\centering\small\setlength{\tabcolsep}{3.5pt}}
  {\end{table}}
\newenvironment{appcolfigure}
  {\begin{figure}[!htbp]\centering}
  {\end{figure}}

\newcommand{\method}{GOPR-CAD}
\title{\method{}: Graph-of-Operations Progressive Reasoning for CAD Generation}

\author{
    Yitian Pang\equalcontrib\textsuperscript{\rm 1},
    Yifan Li\equalcontrib\textsuperscript{\rm 2},
    Xiaowei Yue\textsuperscript{\rm 2}\corresponding,
    Li Zheng\textsuperscript{\rm 2}
}
\affiliations{
    \textsuperscript{\rm 1}Department of Automation, Tsinghua University\\
    \textsuperscript{\rm 2}Department of Industrial Engineering, Tsinghua University\\
    30 Shuangqing Road, Haidian District, Beijing 100084, China\\
    pangyt23@mails.tsinghua.edu.cn, lyf21@mails.tsinghua.edu.cn\\
    yuex@tsinghua.edu.cn, lzheng@mail.tsinghua.edu.cn
}

\begin{document}

\maketitle

\begin{abstract}
Generating accurate CAD models requires substantial
engineering expertise and careful control of geometry,
topology, and mechanical constraints. Text-to-CAD aims to
simplify this process, but current LLM-based methods often
generate CAD programs directly. Language models remain
unreliable when producing long code while coordinating
coupled geometric, topological, and dimensional constraints.
Consequently, executable programs may still produce
structurally incorrect models, which execution rate and
rendered appearance may fail to reveal.
To address these limitations, we propose
\textbf{GOPR-CAD}, an agentic Text-to-CAD pipeline based on
\textbf{Graph-of-Operations Progressive Reasoning}. A
language model acts as a decision agent over encapsulated CAD
operations and progressively constructs an executable
operation graph, which a deterministic compiler converts into
the final CAD model. This separation requires no additional
training and supports different language models.
To evaluate structural fidelity, we construct MHCAD, a
benchmark of 300 standard mechanical part requirements
derived from \textit{Machinery's Handbook}. Their explicit
structures enable reference-free STEP/B-rep evaluation.
GOPR-CAD achieves 82.7\%
Structural Yield on MHCAD, and 87.2\% Structural Yield on an
external CADPrompt subset.
\end{abstract}

\section{Introduction}

Computer-aided design (CAD) models are fundamental to
mechanical design, manufacturing, and engineering analysis.
Compared with rendered images and polygonal meshes, CAD
models encode exact geometry within an explicit topological
structure and retain an editable construction history. This
representation supports reliable downstream processing
throughout the engineering workflow. A useful CAD model must
therefore go beyond a recognizable shape. It must be
geometrically precise, topologically valid, and editable as
an engineering artifact.

Recent research has increasingly automated CAD construction
from geometric observations and large CAD collections.
Reconstruction methods recover analytic geometry from point
clouds, then assemble it into a valid topological structure
\citep{liu2024point2cad}. Sequence models learn ordered
construction histories built from sketches followed by
feature operations
\citep{wu2021deepcad,willis2021fusion,xu2022skexgen}.
Other methods directly synthesize native boundary
representations or parametric surfaces
\citep{xu2024brepgen,usama2026nurbgen}. These advances have
improved geometric recovery and made editable CAD generation
more practical. Most still require geometric input or prior
CAD examples, leaving users to translate design intent into
specialized representations. A more accessible interface
should allow the intended model to be described directly in
natural language.

Text-to-CAD provides such an interface by translating natural
language requirements into editable CAD models. It simplifies
the modeling process by allowing users to express design
intent without mastering specialized CAD commands. Existing
methods map language to parametric command sequences or
structured CAD representations
\citep{khan2024text2cad,wangtext,wang2025cad,
usama2026nurbgen}. More recent work connects large language
models (LLMs) with scriptable interfaces such as CadQuery,
using executable programs to bridge language understanding
and parametric modeling
\citep{xie2025text,guan2026cad,
alrashedy2025generating}. These programs can be executed by
mature CAD kernels and remain available for subsequent
editing. Together, these advances make natural language a
practical interface for creating parametric CAD models.

\begin{figure*}[t]
    \centering
    \includegraphics[width=0.9\textwidth]{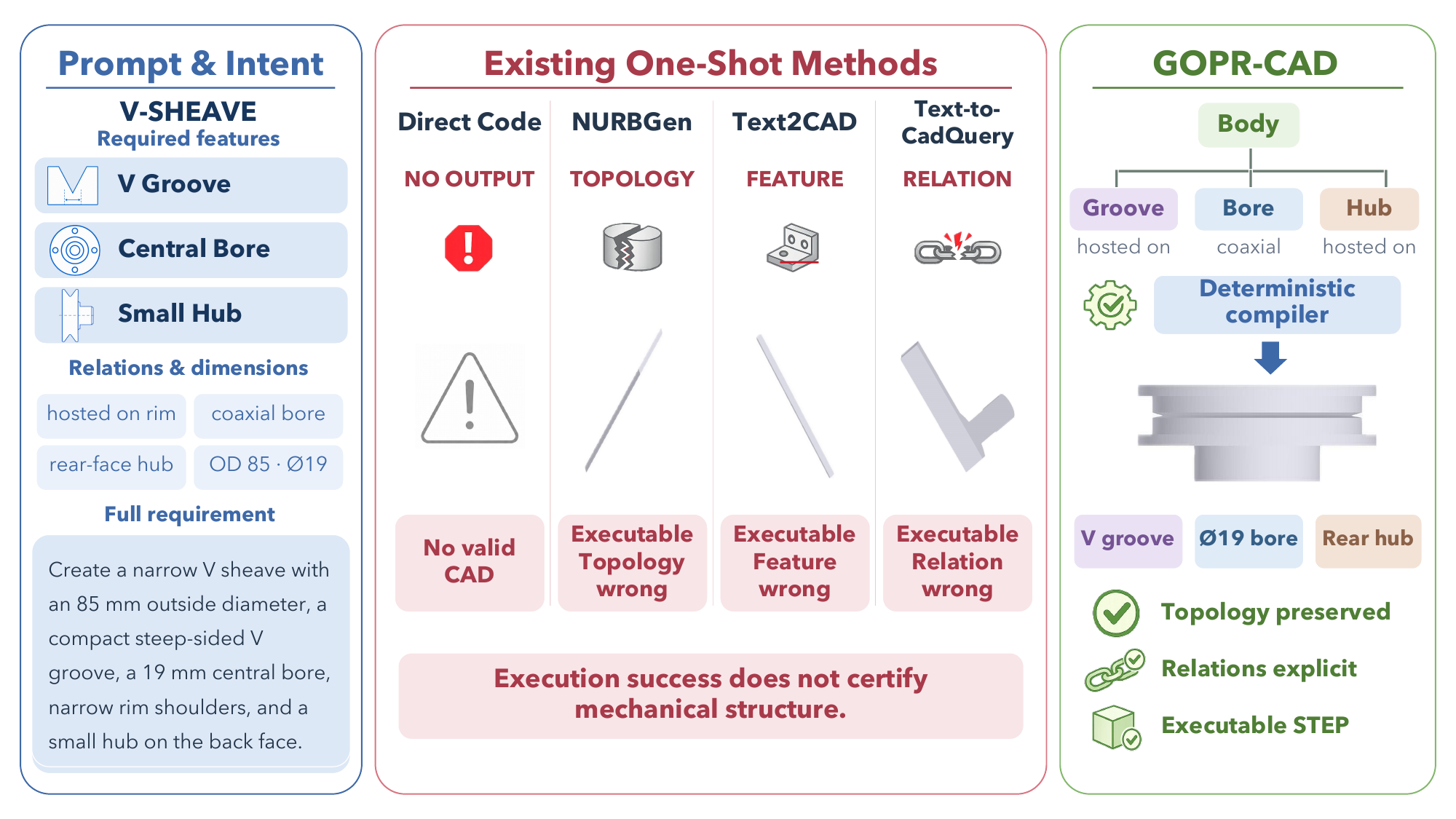}
    \caption{
    The gap between generating an executable CAD model and
    realizing the mechanical structure specified by the
    requirement. Existing methods may fail to produce a valid
    model or may return a valid solid that omits required
    features and relations. \method{} constructs an explicit
    graph of executable operations before deterministic CAD
    realization.
    }
    \label{fig:motivation}
\end{figure*}

However, direct program generation remains unreliable for
complex mechanical parts, as an LLM must preserve geometric,
topological, and dimensional constraints throughout a long
sequence of dependent instructions. Errors introduced early
can propagate through later construction, while the CAD
kernel may still accept the resulting program and export a
valid solid that does not satisfy the requirement. Existing
evaluation protocols often fail to expose this gap.
Execution-based metrics verify artifact validity, while
rendered views may miss errors in internal structure.
Comparison with a single reference is also restrictive, as
the same requirement can admit different valid
constructions. Reliable Text-to-CAD therefore requires a
more dependable construction process and an evaluation
protocol that measures structural fidelity directly on the
generated CAD model.

To address these challenges, we propose
\textbf{\method{}}, an agentic Text-to-CAD pipeline based on
\textbf{Graph-of-Operations Progressive Reasoning}. A language model acts as a decision agent over encapsulated
CAD operations and progressively constructs an executable
operation graph. A deterministic compiler then realizes the
graph as the final CAD model. This separation lets the
language model focus on construction decisions while the CAD
operation library handles geometric realization. It also
decouples the agent backbone from the CAD system, allowing
different language models to be substituted without
additional training or modifications to the remaining
pipeline.

We further construct \textbf{MHCAD} to evaluate the
structural failures that conventional metrics overlook. The
benchmark contains 300 requirements for standard mechanical
parts derived from \textit{Machinery's Handbook}. Standard
parts specify precise features, topology, and dimensional
relations, making them suitable for testing whether a
generated model preserves the intended mechanical structure.
MHCAD derives structural checks from each requirement and
applies them to reimported STEP/B-rep models without assuming
a unique reference construction. Structural Fidelity (SF)
measures structural correctness among valid B-reps, whereas
Structural Yield (SY) measures correct structural delivery
across the full benchmark.

Our contributions are summarized as follows:
\begin{itemize}
    \item We propose \textbf{\method{}}, a training-free
    agentic method that formulates Text-to-CAD as
    \textbf{Graph-of-Operations Progressive Reasoning}. It
    replaces direct CAD program generation with progressive
    decisions over encapsulated operations. By decoupling the
    language model from deterministic CAD realization,
    \method{} allows agent backbones to be replaced without
    retraining or changes to the CAD system.

    \item We construct \textbf{MHCAD}, a benchmark of 300
    standard mechanical part requirements derived from
    \textit{Machinery's Handbook}, together with a
    reference-free STEP/B-rep evaluation protocol based on
    Structural Fidelity and Structural Yield. The protocol
    directly evaluates mechanical structure and reveals
    errors that execution success or rendered appearance may
    overlook.

    \item Extensive experiments show that \method{} achieves
    the best structural performance among all evaluated
    methods, reaching 89.8\% SF and 82.7\% SY on MHCAD and
    87.2\% SY on an external CADPrompt subset. Its advantages
    remain consistent across different language model
    families.
\end{itemize}

\section{Related Work}

\subsection{CAD Generation and Reconstruction}

Methods for CAD generation have developed through
construction sequence modeling, geometric reconstruction,
and direct generation of CAD representations. Large datasets
such as ABC and Fusion 360 Gallery provide geometry,
topology, and construction histories for learning CAD design
distributions~\citep{koch2019abc,willis2021fusion}. DeepCAD
and SkexGen represent CAD models as ordered sequences of
sketches and feature operations
\citep{wu2021deepcad,xu2022skexgen}. Reconstruction methods
recover analytic CAD structures from geometric observations;
Point2CAD, for example, reconstructs CAD models from point
clouds~\citep{liu2024point2cad}. Other approaches directly
generate boundary representations or parametric surfaces,
extending generative modeling from command sequences to
native CAD geometry
\citep{xu2024brepgen,usama2026nurbgen}.

These studies have substantially advanced the generation of
valid geometry, topology, and editable construction
histories. Their inputs and intermediate representations are
usually defined by geometric observations, existing CAD
models, or learned command distributions. Natural language
requirements introduce a different task: the system must
infer which mechanical features should exist, how they are
related, and which dimensions govern their construction.
\method{} addresses this setting through an explicit graph of
executable CAD operations, allowing construction decisions
to be organized before geometric realization.

\subsection{CAD Generation from Language}

Text-to-CAD makes parametric modeling accessible through
natural language. Text2CAD translates language requirements
into sequential CAD commands~\citep{khan2024text2cad}.
CADFusion incorporates feedback from rendered views into
language guided CAD generation~\citep{wangtext}, while
CAD-GPT studies multimodal CAD sequence synthesis with
spatial reasoning~\citep{wang2025cad}. NURBGen generates
parametric surface representations from text through NURBS
parameters~\citep{usama2026nurbgen}. These approaches connect
language descriptions with structured CAD representations
and support editable geometric models.

Another line of work uses large language models with
scriptable CAD interfaces such as CadQuery
\citep{xie2025text,guan2026cad,
alrashedy2025generating}. The language model produces a CAD
program that is executed by an existing geometric kernel.
Graph-CAD further introduces a hierarchical graph over parts
and geometric primitives, followed by an action plan and
executable Blender code~\citep{gonglearning}. Its graph
captures compositional structure and geometric constraints
for object generation.

Linear command sequences and complete CAD programs often
encode feature selection, mechanical relations, and
parameter dependencies only implicitly. A program can
execute while omitting a required feature, assigning it to
an incorrect host, or violating a specified layout.
Graph-CAD makes component structure explicit, but focuses on
compositional geometry and Blender based realization.
\method{} uses executable CAD operations as the units of
decision and develops their relations and parameters in an
explicit graph. The language model determines the design
structure, while a deterministic compiler performs CAD
realization.

\subsection{LLM Agents and Structured Planning}

Large language models are increasingly used as agents that
reason over intermediate states and interact with external
tools. ReAct combines reasoning traces with environment
actions~\citep{yao2023react}, while Toolformer studies how
language models can invoke external tools
\citep{schick2023toolformer}. HuggingGPT coordinates
specialized models through language instructions
\citep{shen2023hugginggpt}, and AutoGen supports structured
interaction among language model agents
\citep{wu2024autogen}. Reflexion introduces feedback through
verbal reflection~\citep{shinn2023reflexion}, while Voyager
uses an executable skill library to support continued
decision making in an interactive environment
\citep{wang2023voyager}. These studies show that a language
model can serve as a decision maker over an external action
space instead of acting only as a text generator.

Structured reasoning methods organize intermediate decisions
beyond a single linear chain. Tree of Thoughts explores
multiple reasoning branches~\citep{yao2023tree}, and Graph of
Thoughts represents reasoning units and their dependencies as
a graph~\citep{besta2024graph}. In these methods, the
structured state primarily supports reasoning toward a
textual answer. GOPR-CAD gives the graph a direct engineering
role. Its nodes correspond to executable CAD operations, and
its edges record mechanical or construction relations. Each
agent decision therefore contributes directly to the design
that is later realized by the CAD system.

\subsection{CAD Benchmarks and Evaluation of Mechanical Structure}

CAD generation is commonly evaluated through file validity,
command accuracy, geometric similarity, visual quality, or
program execution. DeepCAD evaluates CAD command sequences
\citep{wu2021deepcad}, while Text2CAD combines measurements
of parametric precision, geometric accuracy, and rendered
quality~\citep{khan2024text2cad}. Recent benchmarks extend
evaluation to parametric CAD driven by language models,
multimodal CAD programs, and designs with manufacturing or
assembly requirements
\citep{wang2026text2cad,doris2026cadbench,
dong2026muse}. CADCodeVerify studies vision and language
verification for CAD generation and introduces CADPrompt as
a collection of requirements written by domain experts
\citep{alrashedy2025generating}.

These benchmarks provide important measurements of
executability, geometric quality, and correspondence with
reference models. Mechanical requirements also contain
structural conditions that may not be captured by these
measurements. Similar rendered appearances can conceal
differences in feature ownership, internal topology,
alignment, containment, or repeated layout. Reference
geometry can also be restrictive because a requirement may
admit several valid construction strategies. MHCAD therefore
evaluates prompt derived structural conditions on reimported
B-reps.

\section{Method}

\begin{figure*}[t]
    \centering
    \includegraphics[width=\textwidth]{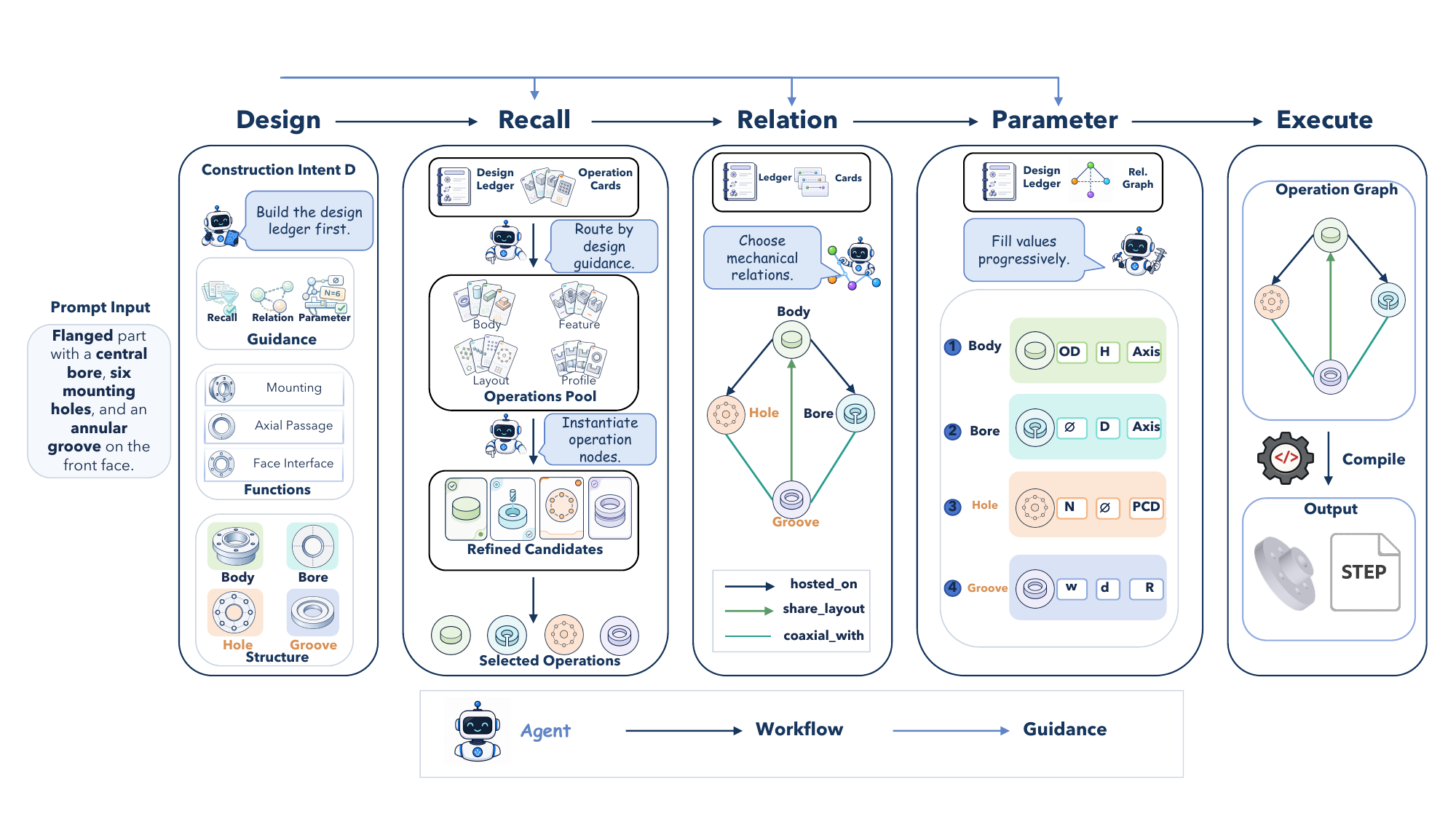}
    \caption{
    Overview of \method{}. A language requirement is
    progressively transformed into construction intent,
    executable operation nodes, typed mechanical relations,
    and complete parameter bindings. The completed operation
    graph is compiled and executed through a deterministic
    CAD realization process.
    }
    \label{fig:pipeline}
\end{figure*}

\subsection{Overview}

\method{} formulates Text-to-CAD as
\textbf{Graph-of-Operations Progressive Reasoning}. Given a
natural language requirement $x$, a language model acts as a
decision agent over a fixed library of executable CAD
operations. The agent develops an explicit operation graph,
while the geometric implementation of each operation and the
final compilation process remain deterministic.

Let
\[
\mathcal{O}=\{o_j\}_{j=1}^{M}
\]
denote the operation library. Each operation is associated
with a contract
\[
o_j=\left(d_j,\mathcal{R}_j,\Theta_j,\kappa_j\right),
\]
where $d_j$ describes its mechanical role,
$\mathcal{R}_j$ specifies the relations in which it may
participate, $\Theta_j$ defines its required parameter
interface, and $\kappa_j$ binds the operation to a
deterministic CAD implementation. The agent selects and organizes
operations without generating their internal geometry.

The graph is completed through four successive
transformations:
\[
x
\xrightarrow{\Phi_D^{\theta}}
D
\xrightarrow{\Phi_O^{\theta}}
G_O
\xrightarrow{\Phi_R^{\theta}}
G_R
\xrightarrow{\Phi_P}
G_P .
\]
Here $D$ represents construction intent, $G_O$ contains the
selected operation nodes, $G_R$ adds typed mechanical
relations, and $G_P$ contains complete parameter bindings.
The superscript $\theta$ denotes decisions made by the
language model. Parameter completion $\Phi_P$ follows
deterministic rules defined by the operation contracts. Each
stage introduces one type of commitment: intent, operations,
relations, and parameters.

The first three stages form a constrained agent process. Let
$H_t$ denote the current structured state and define
\[
s_t=(x,H_t,\mathcal{O}).
\]
At stage $t$, the agent selects a structured decision and
updates the state:
\[
a_t\sim
\pi_{\theta}\!\left(
\cdot\mid s_t,\mathcal{A}_t(s_t)
\right),
\qquad
H_{t+1}=\mathcal{U}_t(H_t,a_t),
\]
where $\mathcal{A}_t(s_t)$ is the set of decisions allowed by
the current stage and the operation contracts.

The contracts restrict each stage to choices supported by the
operation interfaces, removing type-incompatible or
non-executable decisions before CAD realization. This
interface constraint does not guarantee that the agent
selects a prompt-consistent design; its practical effect is
evaluated through the typed-interface ablations. Because the
contracts and compiler are independent of the policy, the
agent backbone can be replaced without retraining or changes
to the CAD system.

\subsection{Design: Representing Construction Intent}

The Design stage establishes a persistent representation of
what the requirement asks the CAD model to contain:
\[
D=\Phi_D^{\theta}(x),
\qquad
D=(C,\Lambda).
\]
Here
\[
C=\{c_i\}_{i=1}^{n}
\]
is a set of construction items, and $\Lambda$ stores the
evidence associated with them. The evidence includes
mechanical roles, feature counts, layout cues, visible
structures, relation cues, and parameter information stated
in the requirement.

The resulting representation provides a common semantic
reference for all later stages. Recall uses it to select
operations, Relation uses it to recover mechanical
dependencies, and Parameters uses it to resolve values and
constraints. Maintaining this shared reference prevents the
later stages from relying on separate interpretations of the
original requirement. The Design state records the required
construction structure while leaving executable realization
to subsequent stages.

\subsection{Recall: Grounding Intent in Executable Operations}

Recall maps each construction item to operations from the
fixed library. The library is organized into capability
families
\[
\mathcal{F}=\{F_k\}_{k=1}^{K},
\]
where each family groups operations that serve related
mechanical roles. For each construction item $c_i$, Recall
first selects a capability family:
\[
F_i=
R_{\mathrm{coarse}}\!\left(
 c_i,\Lambda_i,\mathcal{F}
\right).
\]
It then selects one or more concrete operations from that
family:
\[
V_i=
R_{\mathrm{fine}}\!\left(
 c_i,\Lambda_i,\mathcal{O}(F_i)
\right),
\]
where $\mathcal{O}(F_i)$ is the subset of operations in
$F_i$.

The complete node set and operation graph are
\[
V_O=\bigcup_{i=1}^{n}V_i,
\qquad
G_O=(V_O,\alpha),
\]
where
\[
\alpha:V_O\rightarrow C
\]
links every selected operation to the construction item that
motivated it. This evidence mapping makes operation selection
traceable to the requirement. The two step selection process
reduces the number of operations considered for each item
while retaining composition across mechanical roles. Since
every node is selected from $\mathcal{O}$, it already has a
known parameter interface and an implementation in the CAD
system. The language model therefore selects executable
building blocks without inventing geometric procedures.

\subsection{Relation: Composing the Operation Graph}

A set of valid operations does not yet define a coherent
mechanical model. Their ownership, alignment, containment,
layout, and construction dependencies must be explicit.
Relation augments $G_O$ with typed edges:
\[
G_R=(V_O,E_R,\alpha),
\]
where
\[
E_R\subseteq V_O\times\mathcal{T}_R\times V_O.
\]

We use the relation vocabulary
\[
\mathcal{T}_R=
\{\mathrm{host},\mathrm{align},\mathrm{contain},
\mathrm{refine},\mathrm{pattern},\mathrm{depend}\}.
\]
The relations $\mathrm{host}$ and $\mathrm{contain}$ describe
feature ownership and enclosure. The relations
$\mathrm{align}$ and $\mathrm{pattern}$ express shared axes,
orientations, and repeated layouts. The relations
$\mathrm{refine}$ and $\mathrm{depend}$ represent
construction and parameter dependencies.

Relation decisions are restricted by the operation
contracts. For an edge
\[
e_{uv}=(v_u,\tau_{uv},v_v),
\]
the selected type must satisfy
\[
\tau_{uv}\in\Gamma(o_u,o_v),
\]
where $\Gamma(o_u,o_v)$ is the set of relations supported by
the contracts of the two operations. Feature ownership,
shared axes, containment, repeated layout, and parameter
dependencies are therefore available in the graph before CAD
execution.

\subsection{Parameters: Completing Operation Interfaces}

The Parameters stage resolves the interfaces of the selected
operations. Values may come directly from the requirement,
be propagated through graph relations, or be derived through
rules defined for each operation:
\[
P=P_{\mathrm{req}}\cup P_{\mathrm{rel}}\cup P_{\mathrm{rule}}.
\]
Here $P_{\mathrm{req}}$ contains dimensions and counts stated
in the requirement, $P_{\mathrm{rel}}$ contains values shared
through alignment, hosting, and pattern relations, and
$P_{\mathrm{rule}}$ contains values determined by operation
constraints and valid parameter domains.

Parameter completion follows the dependency structure of the
graph. Let $V^{(\ell)}$ contain the operations assigned to
dependency level $\ell$. Early levels usually contain primary
bodies and reference geometry, while later levels contain
hosted features, repeated structures, and refinements.
Parameters are accumulated as
\[
P=\bigcup_{\ell=0}^{L}P^{(\ell)},
\]
where
\[
P^{(\ell)}=
\Phi_P^{(\ell)}\!\left(
D,G_R,V^{(\ell)},P^{(<\ell)}
\right),
\qquad
P^{(<\ell)}=\bigcup_{j<\ell}P^{(j)}.
\]
This ordering allows a later operation to use frames, axes,
dimensions, and counts that have already been resolved for
its dependencies.

The completed graph is
\[
G_P=(V_O,E_R,P,\alpha).
\]
It is ready for execution when every required operation
parameter is available:
\[
\forall v\in V_O,
\qquad
\operatorname{Req}(v)\subseteq\operatorname{Dom}(P_v),
\]
and every edge satisfies the relation contracts of its
incident operations. Unresolved values are retained as typed
parameter errors so incomplete commitments remain visible.
Once $G_P$ is complete, a deterministic compiler converts the
graph into an executable CAD operation plan:
\[
\Pi=\mathcal{C}(G_P),
\qquad
y=\mathcal{K}(\Pi),
\]
where $\mathcal{C}$ is the compiler, $\mathcal{K}$ is the CAD
kernel, and $y$ is the resulting CAD model. No further
language model decisions are introduced during realization.

\section{Experiments}

\subsection{Experimental Setup}

We evaluate \method{} on two prompt sources. \textbf{MHCAD}
contains 300 requirements for single standard mechanical
parts and 342 applicable structural checks. The external
\textbf{CADPrompt subset} contains 30 single part cases with
47 applicable checks, selected independently of model
outputs. The second source tests whether the structural
advantage transfers beyond prompts constructed for MHCAD.
Detailed distributions, annotation rules, and subset
construction are provided in the supplementary material.

We report three geometry based metrics. \textbf{STEP}
measures the percentage of prompts for which a STEP file is
produced. Let $v_i\in\{0,1\}$ indicate whether case $i$
yields a valid B-rep after STEP reimport, and let
$\mathcal{A}_i$ denote its applicable structural checks.
Structural Fidelity is
\[
\mathrm{SF}
=
\frac{
\sum_{i:v_i=1}\sum_{j\in\mathcal{A}_i}
\mathbf{1}[\mathrm{pass}_{ij}]
}{
\sum_{i:v_i=1}|\mathcal{A}_i|
}
\times 100,
\]
while Structural Yield is
\[
\mathrm{SY}
=
\frac{
\sum_i\sum_{j\in\mathcal{A}_i}
\mathbf{1}[v_i=1\land\mathrm{pass}_{ij}]
}{
\sum_i|\mathcal{A}_i|
}
\times 100.
\]
SF measures structural correctness among valid B-reps. SY
measures correct structural delivery over the complete
benchmark and gives no structural credit to missing or
invalid outputs. The checks cover bore presence, body
structure, coaxiality, mounting holes, repeated circular
patterns, and rolling element count. Complete definitions and
applicability rules are reported in the supplementary material.

We compare against Text2CAD, Text-to-CadQuery, CADFusion,
NURBGen, Qwen3-Coder-30B Direct, and DeepSeek-V4-Flash
Direct. The baselines cover CAD command sequences,
parametric surfaces, and direct generation through language
models. All methods use the same STEP reimport gate,
structural records, checker definitions, and frozen checker
settings.

Unless otherwise stated, \method{} uses Qwen3.5-9B as the
agent backbone~\citep{yang2025qwen3}. The operation library,
compiler, and CAD kernel remain fixed across agent backbone
experiments. The experiments were executed on six NVIDIA
H100 GPUs. We estimate uncertainty with 10,000 bootstrap
replicates clustered by prompt, so every check attached to a
sampled prompt is resampled together. A destructive mutation
test removes the topology targeted by each checker while
preserving a valid CAD artifact; the frozen checkers detect
all 86 valid mutations.

\subsection{Main Results}

\begin{table}[t]
\centering
{\small
\setlength{\tabcolsep}{3.5pt}%
\begin{tabular}{@{}lrrr@{}}
\toprule
Method & STEP & SF & SY \\
\midrule
\method{} (Ours) & 90.0 & \textbf{89.8} & \textbf{82.7} \\
Text2CAD & \underline{94.0} & 40.8 & 37.4 \\
Text-to-CadQuery & 79.0 & 39.4 & 32.7 \\
CADFusion & 53.0 & 64.0 & 30.7 \\
NURBGen & \textbf{97.3} & 21.8 & 20.8 \\
Qwen3-Coder Direct & 43.7 & \underline{66.9} & 28.4 \\
DeepSeek-V4 Direct & 39.0 & 66.7 & 24.6 \\
\bottomrule
\end{tabular}
}
\caption{
Main structural results on MHCAD (300 prompts; 342 applicable
checks). SF is conditioned on valid B-reps; SY measures
structural delivery over all applicable checks. Complete counts
and confidence intervals are in the supplement.
}
\label{tab:main_results}
\end{table}

\begin{figure}[t]
    \centering
    \includegraphics[width=\columnwidth]{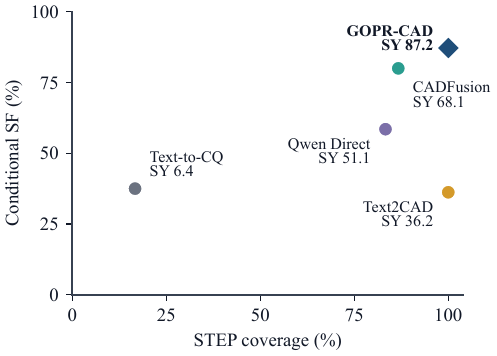}
    \caption{
    Coverage--fidelity tradeoff on the external CADPrompt subset
    (30 prompts; 47 checks). Coordinates show STEP coverage and
    conditional SF; labels report SY. Only methods with reported
    results are shown; exact counts are in the supplement.
    }
    \label{fig:cadprompt_subset}
\end{figure}

Table~\ref{tab:main_results} shows that \method{} reaches
89.8 SF and 82.7 SY on MHCAD. Text2CAD and NURBGen
produce STEP files for 94.0\% and 97.3\% of prompts, but
their SY values are 37.4 and 20.8: file coverage does not imply
delivery of the requested mechanical structure.

Direct language model baselines obtain stronger conditional
SF on surviving outputs than several sequence and surface
methods, but low STEP coverage keeps their MHCAD SY below
30.0. \method{} instead combines high conditional quality
with broad delivery, as reflected by its smaller SF--SY gap.

Figure~\ref{fig:cadprompt_subset} shows the same pattern on
the external subset. Without changing the operation library or
protocol, \method{} reaches 87.2 SY; CADFusion is the
strongest baseline at 68.1. On MHCAD, every bootstrap 95\%
interval for the pairwise SY margin remains above zero.

\subsection{Analysis of Structural Correctness}

\begin{table*}[t]
\centering
{\small
\setlength{\tabcolsep}{5.0pt}%
\begin{tabular}{@{}lrrrrrr@{}}
\toprule
Method & Bore & Body & Coax. Axis & Mount Holes & Circ. Pattern & Rolling Count \\
\midrule
\textbf{\method{} (Ours)} & 99.0 & 75.0 & \textbf{82.5} & \textbf{100.0} & \textbf{96.4} & \textbf{94.7} \\
Text2CAD & 57.4 & 0.0 & 39.1 & 25.0 & 15.4 & 11.8 \\
Text-to-CadQuery & 59.8 & 25.0 & 25.2 & 60.0 & 44.0 & 40.0 \\
CADFusion & \textbf{100.0} & 25.0 & 48.4 & \textbf{100.0} & 55.6 & 50.0 \\
NURBGen & 26.5 & 0.0 & 25.1 & 20.0 & 3.6 & 0.0 \\
Qwen3-Coder Direct & 88.6 & \textbf{100.0} & 58.3 & \textbf{100.0} & 50.0 & 45.5 \\
DeepSeek-V4 Direct & 90.0 & \textbf{100.0} & 44.6 & \textbf{100.0} & 90.9 & 80.0 \\
\bottomrule
\end{tabular}
}
\caption{
Structural Fidelity (\%) for the six validated check types on
MHCAD. Exact numerators and valid-B-rep denominators are
reported in the supplement.
}
\label{tab:structural_checks}
\end{table*}

\begin{figure*}[t]
    \centering
    \includegraphics[width=\textwidth]{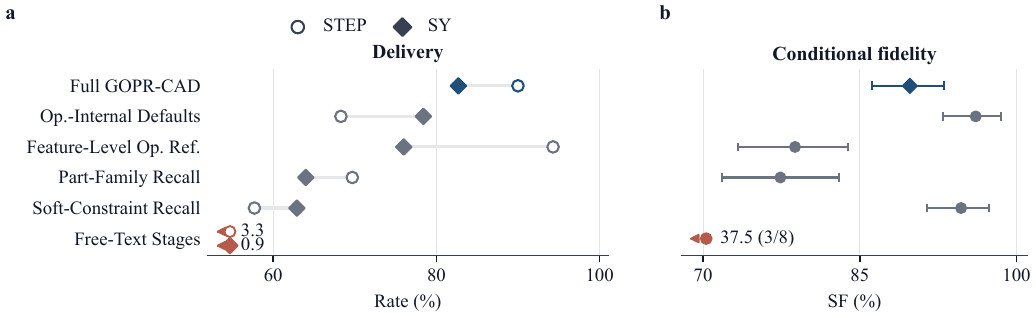}
    \caption{
    Design ablations on MHCAD. \textbf{(a)} STEP coverage and
    Structural Yield. \textbf{(b)} Conditional Structural Fidelity;
    typed variants show 95\% bootstrap intervals. Left-pointing
    markers denote off-scale Free-Text Stages point estimates
    (SF is computed over 3/8 surviving checks and has no interval).
    }
    \label{fig:main_analysis}
\end{figure*}

Table~\ref{tab:structural_checks} localizes the main result.
\method{} exceeds 94\% on bore, mounting-hole, circular-pattern,
and rolling-count checks, and reaches 82.5\% on coaxial axes.
Its remaining errors concentrate on body realization and
relation-sensitive placement. Some baselines reach 100\% on
sparse checks, but none maintains comparable performance
across all six; the supplement reports the exact denominators.

\subsection{Ablation Study}

Figure~\ref{fig:main_analysis} first weakens choices within typed
stage interfaces. Moving defaults inside operations and using
coarser feature references reduce SY by 4.3 and 6.7 points;
Part-Family Template Recall and Soft-Constraint Recall lose 18.7
and 19.8 points. Free-Text Stages tests the typed executable
boundary under the same four-stage workflow. It retains the
stage order, stage-specific prompts, closed operation
documentation, agent backbone, and decoding configuration, but
replaces typed executable state and deterministic compilation
with prose intermediates followed by direct CadQuery realization.
All stages return outputs for every prompt, yet STEP falls to
3.3\% and SY to 0.9\%, with 251 of 300 failures occurring
during code execution. Under this matched workflow, prose
intermediates followed by direct code realization do not
provide the delivery reliability of the typed executable
pipeline. Within the typed pipeline, closed compositional
operation grounding remains the most important design factor.
Full controls and failure attribution are reported in the
supplementary material.

\begin{center}
\begin{minipage}{\columnwidth}
    \centering
    \includegraphics[width=\columnwidth]{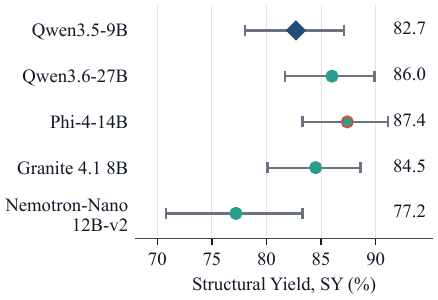}
    \captionof{figure}{
    Structural Yield across agent backbones on MHCAD. Bars show
    95\% bootstrap intervals; the diamond marks the default.
    Exact counts and conditional SF are in the supplement.
    }
    \label{fig:agent_backbones}
\end{minipage}
\end{center}

\subsection{Generalization Across Agent Backbones}

Figure~\ref{fig:agent_backbones} shows SY from 77.2 to 87.4
across five backbones from four model families. Phi-4-14B is
highest at 87.4, while Granite 4.1 8B reaches 84.5; performance
is not monotonic in parameter count. Different model families
can therefore serve as the decision agent without changing the
CAD implementation.

\section{Conclusion}

We introduced \method{}, which separates language-model construction
decisions from typed operations and deterministic compilation. It
achieves 89.8\% SF and 82.7\% SY on MHCAD and 87.2\% SY on CADPrompt.
The 0.9\% SY of Free-Text Stages shows that, under the
matched staged workflow, prose intermediates followed by
direct code realization do not match the delivery reliability
of the typed executable pipeline.

Future work will extend the operation graph to complex assemblies with
mating constraints, motion relations, cross-part dependencies, and
assembly-aware compilation. It will also use graph validity, compiler
diagnostics, SF, SY, and checker outcomes as reinforcement-learning
feedback, moving the training-free agent toward a learned policy while
preserving deterministic geometric execution.

\clearpage
\setlength{\bibsep}{1pt}
\bibliography{references}

\clearpage
\appendix
\setcounter{secnumdepth}{2}
\setcounter{section}{0}
\setcounter{subsection}{0}
\setcounter{table}{0}
\setcounter{figure}{0}
\setcounter{algorithm}{0}

\makeatletter
\@addtoreset{table}{section}
\@addtoreset{figure}{section}
\@addtoreset{algorithm}{section}
\makeatother

\renewcommand{\thesection}{\Alph{section}}
\renewcommand{\thetable}{\Alph{section}\arabic{table}}
\renewcommand{\thefigure}{\Alph{section}\arabic{figure}}
\renewcommand{\thealgorithm}{\Alph{section}\arabic{algorithm}}



\section{MHCAD Benchmark Construction}
\label{app:mhcad}

MHCAD contains 300 single-part requirements derived from
\textit{Machinery's Handbook}. This section documents its
category coverage, annotation protocol, and representative
structural records.

\subsection{Category Distribution}

\begin{appcoltable}
\begin{tabularx}{\columnwidth}{@{}Yrr@{}}
\toprule
Category or mechanical family
& Prompts
& Percentage \\
\midrule
Bearings, bushings, and rolling elements
& 79
& 26.3\% \\

Fasteners and washers
& 73
& 24.3\% \\

Power transmission
& 44
& 14.7\% \\

Shafts, keys, retainers, and seals
& 35
& 11.7\% \\

Structural sections and stock
& 23
& 7.7\% \\

Springs and formed wire
& 21
& 7.0\% \\

Couplings and connectors
& 18
& 6.0\% \\

Other standard-part controls
& 5
& 1.7\% \\

Cams, followers, and linkages
& 2
& 0.7\% \\
\bottomrule
\end{tabularx}
\caption{MHCAD category distribution (300 prompts).}
\label{tab:app_mhcad_categories}
\end{appcoltable}

\subsection{Prompt Complexity and Structural Coverage}

MHCAD contains 342 applicable checks across 196 prompts. The
remaining 104 prompts are retained in the STEP and validity
denominators and have no structural-check entries.
Figure~\ref{fig:app_mhcad_complexity} summarizes prompt
complexity and check coverage.

\begin{appcolfigure}
\includegraphics[
  width=0.96\columnwidth,
  trim=0 0 254.35 0,
  clip
]{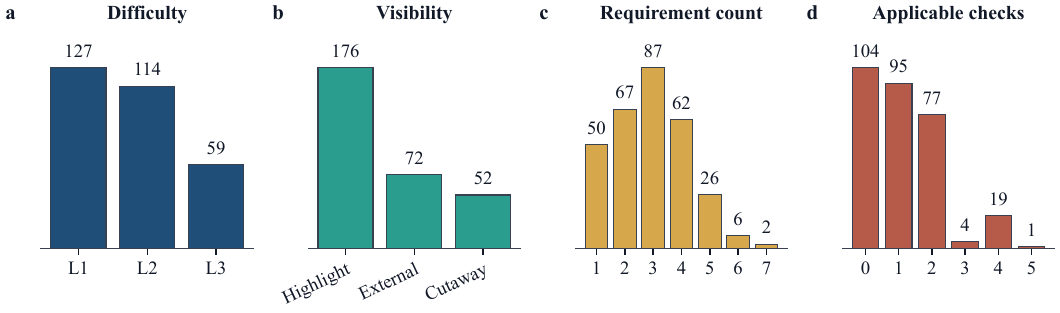}
\par\vspace{-1mm}
\includegraphics[
  width=0.96\columnwidth,
  trim=254.35 0 0 0,
  clip
]{fig/app_mhcad_complexity.pdf}
\caption{Prompt difficulty, visibility, requirement
count, and structural check coverage in MHCAD.}
\label{fig:app_mhcad_complexity}
\end{appcolfigure}

\FloatBarrier

\subsection{Structural Annotation Protocol}

Before evaluation, benchmark metadata define each record's
applicable checks, expected counts or relations, and any
requirements outside the deterministic check set.

\begin{center}
\begin{minipage}{\columnwidth}
\centering
\small
\setlength{\tabcolsep}{3.5pt}
\begin{tabularx}{\columnwidth}{@{}>{\raggedright\arraybackslash}p{0.28\columnwidth}Y@{}}
\toprule
Step & Rule and output \\
\midrule
Requirement analysis
& From the prompt and mechanical family metadata, identify
required features, relations, counts, layouts, and dimensional
cues; record the candidate structural requirements. \\

Applicability filtering
& Include a deterministic STEP/B-rep check only when the prompt
supplies sufficient evidence; record the applicable validated
checks for each case. \\

Deterministic evaluation scope
& For small ridges, sockets, decorative marks, and exact gear
teeth, retain details supported by stable topology extraction
rules. SF and SY include only requirements supported by
validated deterministic checks. \\

Prompts without structural checks
& Retain prompts with no validated check in STEP and validity
evaluation, with no structural denominator entries. Coverage
remains explicit while structural scores rely on validated
checks. \\

Shared protocol
& Apply the same structural records, checker definitions, and
tolerances to outputs from all methods, enabling comparable
evaluation across methods. \\
\bottomrule
\end{tabularx}
\captionof{table}{Structural annotation procedure for MHCAD.}
\label{tab:app_gt_construction}
\end{minipage}
\end{center}

SY retains every applicable check in its denominator,
including checks associated with missing outputs. Prompts
without a validated check remain part of the benchmark and
contribute only to STEP and validity measures.

\subsection{Representative Prompt Annotations}

Table~\ref{tab:app_prompt_examples} shows representative
prompt annotations. The complete prompt collection is part of
the benchmark material.

\begin{center}
\begin{minipage}{\columnwidth}
\centering
\small
\setlength{\tabcolsep}{3.5pt}
\begin{tabularx}{\columnwidth}{@{}p{0.25\columnwidth}Y@{}}
\toprule
Case & Annotation record \\
\midrule
Deep groove radial ball bearing
& \textit{Prompt:} Bearing with outer ring, inner bore, eight
balls, and concave raceway grooves.
\textit{Family:} Bearings.
\textit{Relations/dimensions:} Bore coaxial with outer ring;
balls evenly placed; 62 mm outer ring; 30 mm bore; eight balls.
\textit{Checks:} Bore, Coaxial Axis, Rolling Count, Circular Pattern. \\

Classical V sheave
& \textit{Prompt:} Sheave with one V groove, central bore,
raised hub, and rounded rim edges.
\textit{Family:} Power transmission.
\textit{Relations/dimensions:} Bore coaxial with body; V groove
hosted on rim; 90 mm outside diameter; 20 mm bore.
\textit{Checks:} Bore, Coaxial Axis. \\

O ring in rectangular gland
& \textit{Prompt:} O ring material body, rectangular gland,
center bore, compression cue, and flat shoulders.
\textit{Family:} Seal or gland hardware.
\textit{Relations/dimensions:} O ring seated in gland; bore
passes through block; 6 mm seal section; 52 mm bore.
\textit{Checks:} Body, Coaxial Axis. \\

Two piece journal bearing
& \textit{Prompt:} Split shell halves, central bore, split
line, cap screws, and dowels.
\textit{Family:} Bearings.
\textit{Relations/dimensions:} Shells share a bore; cap screws
cross the joint; 34 mm bore.
\textit{Checks:} Bore, Coaxial Axis, Mount Holes, Circular Pattern. \\
\bottomrule
\end{tabularx}
\captionof{table}{Representative MHCAD prompts and structural annotations.}
\label{tab:app_prompt_examples}
\end{minipage}
\end{center}


\FloatBarrier
\section{STEP/B-rep Structural Evaluation}
\label{app:structural_eval}

\subsection{Validity Gates and Metric Definitions}

Table~\ref{tab:app_validity_gate} defines the CAD validity
gates applied before structural scoring.

\begin{appcoltable}
\begin{tabularx}{\columnwidth}{@{}p{0.25\columnwidth}Y@{}}
\toprule
Gate & Definition and scoring consequence \\
\midrule
STEP found
& A method produces a STEP file; a missing file remains in the
SY denominator with zero structural passes. \\

STEP reimport
& The file loads in the CAD kernel. SF includes successful
reimports; SY retains failures with zero structural passes. \\

B-rep valid
& The imported shape passes the CAD validity test. SF includes
valid B-reps; SY assigns zero structural delivery to invalid cases. \\

Structural checks
& Applicable checks run on valid reimported B-reps; passes
contribute to SF and SY according to their denominators. \\
\bottomrule
\end{tabularx}
\normalsize
\caption{STEP/B-rep validity gates and scoring consequences.}
\label{tab:app_validity_gate}
\end{appcoltable}

For case $i$, let $v_i\in\{0,1\}$ indicate a valid B-rep
after STEP reimport and let $\mathcal{A}_i$ contain its
applicable checks. Structural Fidelity is
\[
\mathrm{SF}
=
\frac{
\sum_{i:v_i=1}\sum_{j\in\mathcal{A}_i}
\mathbf{1}[\mathrm{pass}_{ij}]
}{
\sum_{i:v_i=1}|\mathcal{A}_i|
}
\times 100.
\]
Structural Yield is
\[
\mathrm{SY}
=
\frac{
\sum_i\sum_{j\in\mathcal{A}_i}
\mathbf{1}[v_i=1\land\mathrm{pass}_{ij}]
}{
\sum_i|\mathcal{A}_i|
}
\times 100.
\]
SF is conditional on valid B-reps. SY measures delivery over
the complete benchmark and assigns zero structural passes to
missing STEP, failed reimport, and invalid B-rep outputs.
Prompts without applicable checks remain in STEP and validity
denominators and are omitted from structural denominators.

\subsection{Structural Check Definitions}

\begin{center}
\begin{minipage}{\columnwidth}
\centering
\small
\setlength{\tabcolsep}{3.5pt}
\begin{tabularx}{\columnwidth}{@{}p{0.22\columnwidth}Y@{}}
\toprule
Check & Definition and evaluation contract \\
\midrule
Bore
& A required central or axial bore is a circular opening or
annular void. The check applies to through bores, sleeve
openings, hub bores, and annular openings, and passes when
valid topology contains a consistent central or annular
circular radius group. Its scope is limited to openings
represented by stable topology. \\

Body
& A required separate material body exists instead of only a
cut or groove. The check applies to O rings, retaining rings,
rolling bodies, and inserted members, and passes with a
distinct body or validated material-volume cue. Kernel fusion
is handled through the documented cue. \\

Coaxial Axis
& A required coaxial or concentric relation is realized. The
check applies to shaft, bore, ring, or hub alignment and passes
when functional circular groups share an axis within the
checker tolerance. Its scope is restricted to functional
circular groups. \\

Mount Holes
& Required mounting or bolt holes are present in the expected
count range. The check applies to mounting holes, screw holes,
bolt pockets, dowels, and perforations, and passes when
detected hole cues meet the expected count. Its deterministic
scope covers stable topological openings. \\

Circular Pattern
& A required regular circular or repeated layout is present.
The check applies to bolt circles, repeated rolling elements,
spokes, and webs, and passes when repeated-feature cues provide
sufficient regularity and count evidence. Exact gear-tooth
counts are outside its scope. \\

Rolling Count
& Required rolling elements are present in a plausible count
and layout. The check applies to balls, rollers, needles, and
rolling rows, and passes when rolling-body cues meet the count
evidence without collapsing to a smooth annular substitute.
Cage separators may be fused by the CAD kernel. \\
\bottomrule
\end{tabularx}
\captionof{table}{Definitions of the structural checks used in SF and SY.}
\label{tab:app_structural_checks}
\end{minipage}
\end{center}

\subsection{Checker Validation}

Checker sensitivity is evaluated by removing or replacing the
topology targeted by each checker while preserving a valid
STEP/B-rep artifact.

\begin{center}
\begin{minipage}{\columnwidth}
\centering
\small
\setlength{\tabcolsep}{3pt}%
\begin{tabularx}{\columnwidth}{@{}YrrY@{}}
\toprule
Check & \shortstack{Valid\\mutants} & Detected & Mutation \\
\midrule
Bore & 20 & 20 & replace with an unbored solid cylinder \\
Body & 3 & 3 & replace with a single host solid \\
Coaxial Axis & 20 & 20 & replace with a nonannular box \\
Mount Holes & 5 & 5 & replace with an unperforated plate \\
Circular Pattern & 20 & 20 & replace with an unpatterned solid \\
Rolling Count & 18 & 18 & replace with a smooth host body \\
\midrule
Total & 86 & 86 & -- \\
\bottomrule
\end{tabularx}
\captionof{table}{Checker validation through topology-targeted destructive
mutations that preserve a valid STEP/B-rep artifact.}
\label{tab:app_checker_validation}
\end{minipage}
\end{center}

The checkers detect all 86 targeted topology mutations while
the resulting STEP/B-rep artifacts remain valid.

\subsection{Evaluation Scope and Metric Interpretation}

SF and SY should be interpreted jointly. High STEP with low
SF indicates that files are produced but the required
structure is frequently missing. High SF with low SY indicates
that the surviving valid outputs are strong but coverage is
limited. High values for both metrics indicate that the
annotated structural requirements are delivered over a large
portion of the benchmark.
For example, full \method{} passes 283 of 315 checks on valid
B-reps and 283 of 342 checks over the complete benchmark,
which yields 89.8\% SF and 82.7\% SY.


\FloatBarrier
\section{Additional Results on MHCAD}
\label{app:mhcad_results}

\subsection{Exact Main Results}

Table~\ref{tab:app_main_count} provides the complete counts
behind the compact main paper comparison.

\begin{appcoltable}
\setlength{\tabcolsep}{2.3pt}%
\begin{tabularx}{\columnwidth}{@{}Ycccc@{}}
\toprule
Method & \shortstack{STEP/\\reimport} & \shortstack{B-rep\\valid} & SF & SY \\
\midrule
\method{} (Ours) & 270/270 & 270/300 & 283/315 & 283/342 \\
Text2CAD & 282/282 & 281/300 & 128/314 & 128/342 \\
Text-to-CadQuery & 237/237 & 228/300 & 112/284 & 112/342 \\
CADFusion & 159/159 & 159/300 & 105/164 & 105/342 \\
NURBGen & 292/292 & 287/300 & 71/326 & 71/342 \\
Qwen3-Coder Direct & 131/131 & 128/300 & 97/145 & 97/342 \\
DeepSeek-V4 Direct & 117/117 & 113/300 & 84/126 & 84/342 \\
\bottomrule
\end{tabularx}
\caption{Exact MHCAD results. STEP/reimport gives the number
of files found and successfully reimported; SF and SY give
passed/eligible checks. All artifact denominators are 300.}
\label{tab:app_main_count}
\end{appcoltable}

\subsection{Statistical Analysis}

Uncertainty is estimated with 10,000 bootstrap replicates
clustered by prompt over the 300 prompt identities, using
seed 20260710. Each replicate preserves all checks attached
to a sampled prompt. Table~\ref{tab:app_main_ci} reports the
resulting intervals and paired SY margins.

\begin{appcoltable}
\setlength{\tabcolsep}{2.4pt}%
\begin{tabular}{@{}lcccc@{}}
\toprule
Method & SF & SF 95\% CI & SY & SY 95\% CI \\
\midrule
\method{} & 89.8 & [86.2, 93.1] & 82.7 & [78.0, 87.1] \\
Text2CAD & 40.8 & [34.4, 47.2] & 37.4 & [31.4, 43.7] \\
Text-to-CQ & 39.4 & [32.3, 46.7] & 32.7 & [26.3, 39.2] \\
CADFusion & 64.0 & [57.0, 70.7] & 30.7 & [24.7, 36.9] \\
NURBGen & 21.8 & [15.9, 28.3] & 20.8 & [15.1, 26.9] \\
Qwen Direct & 66.9 & [58.4, 75.4] & 28.4 & [22.1, 34.9] \\
DeepSeek Direct & 66.7 & [57.4, 75.2] & 24.6 & [18.3, 31.3] \\
\bottomrule
\end{tabular}
\par\medskip
\begin{tabularx}{\columnwidth}{@{}Ycc@{}}
\toprule
Comparison & $\Delta$SY & 95\% CI \\
\midrule
\method{}--Text2CAD & 45.3 & [37.9, 52.4] \\
\method{}--Text-to-CadQuery & 50.0 & [42.2, 57.5] \\
\method{}--CADFusion & 52.0 & [44.1, 59.8] \\
\method{}--NURBGen & 62.0 & [54.2, 69.1] \\
\method{}--Qwen3-Coder Direct & 54.4 & [45.7, 62.5] \\
\method{}--DeepSeek-V4 Direct & 58.2 & [49.7, 66.2] \\
\bottomrule
\end{tabularx}
\caption{Bootstrap intervals and pairwise Structural Yield margins
relative to \method{}. Every pairwise interval remains above zero.}
\label{tab:app_main_ci}
\end{appcoltable}

\subsection{Results by Structural Requirement}

Table~\ref{tab:app_per_check_counts} reports the six
structural-check groups with their exact valid-B-rep
denominators.

\begin{appcoltable}
\setlength{\tabcolsep}{1.6pt}%
\begin{tabularx}{\columnwidth}{@{}Yccc@{}}
\toprule
Method & Bore & Body & \shortstack{Coaxial\\Axis} \\
\midrule
\method{} (Ours) & 98/99 (99.0) & 3/4 (75.0) & \textbf{132/160 (82.5)} \\
Text2CAD & 58/101 (57.4) & 0/5 (0.0) & 63/161 (39.1) \\
Text-to-CQ & 55/92 (59.8) & 1/4 (25.0) & 36/143 (25.2) \\
CADFusion & \textbf{49/49 (100.0)} & 1/4 (25.0) & 45/93 (48.4) \\
NURBGen & 27/102 (26.5) & 0/5 (0.0) & 42/167 (25.1) \\
Qwen Direct & 39/44 (88.6) & \textbf{3/3 (100.0)} & 42/72 (58.3) \\
DeepSeek Direct & 36/40 (90.0) & \textbf{2/2 (100.0)} & 29/65 (44.6) \\
\bottomrule
\end{tabularx}
\par\medskip
\begin{tabularx}{\columnwidth}{@{}Yccc@{}}
\toprule
Method & \shortstack{Mount\\Holes} & \shortstack{Circular\\Pattern} & \shortstack{Rolling\\Count} \\
\midrule
\method{} (Ours) & \textbf{5/5 (100.0)} & \textbf{27/28 (96.4)} & \textbf{18/19 (94.7)} \\
Text2CAD & 1/4 (25.0) & 4/26 (15.4) & 2/17 (11.8) \\
Text-to-CQ & 3/5 (60.0) & 11/25 (44.0) & 6/15 (40.0) \\
CADFusion & \textbf{1/1 (100.0)} & 5/9 (55.6) & 4/8 (50.0) \\
NURBGen & 1/5 (20.0) & 1/28 (3.6) & 0/19 (0.0) \\
Qwen Direct & \textbf{1/1 (100.0)} & 7/14 (50.0) & 5/11 (45.5) \\
DeepSeek Direct & \textbf{3/3 (100.0)} & 10/11 (90.9) & 4/5 (80.0) \\
\bottomrule
\end{tabularx}
\caption{MHCAD results for each structural check, reported as
passed/eligible checks with Structural Fidelity in parentheses.
The two blocks cover all six validated checks reported in the
main paper.}
\label{tab:app_per_check_counts}
\end{appcoltable}


\section{Analysis of GOPR-CAD}
\label{app:gopr_analysis}

\subsection{Design Ablations}

The executable design variants isolate parameter state,
operation granularity, compositional recall, and the strength
of operation contracts.

\begin{appcoltable}
\begin{tabularx}{\columnwidth}{@{}Yccc@{}}
\toprule
Variant & STEP & \shortstack{SF [95\% CI]} & \shortstack{SY [95\% CI]} \\
\midrule
Full \method{} & 270/300 & 89.8 [86.2, 93.1] & 82.7 [78.0, 87.1] \\
Operation-Internal Defaults & 205/300 & 96.1 [93.0, 98.5] & 78.4 [72.7, 83.6] \\
Feature-Level Operation Reference & 283/300 & 78.8 [73.3, 83.9] & 76.0 [70.4, 81.3] \\
Part-Family Template Recall & 209/300 & 77.4 [71.8, 83.0] & 64.0 [58.2, 69.7] \\
Soft-Constraint Recall & 173/300 & 94.7 [91.4, 97.4] & 62.9 [55.5, 69.5] \\
Free-Text Stages & 10/300 & 37.5 (3/8) & 0.9 (3/342) \\
\bottomrule
\end{tabularx}
\caption{Complete design-ablation results on MHCAD. The first
five variants retain the typed executable pipeline.
Free-Text Stages preserves the staged workflow but replaces
typed executable state and deterministic compilation with
prose intermediates followed by direct CadQuery realization.
Its point estimates are reported without bootstrap
intervals.}
\label{tab:app_design_ablations}
\end{appcoltable}

Operation-internal defaults increase conditional SF to 96.1
but reduce STEP coverage and SY to 78.4. Coarser feature
references reduce structural quality, while part-family recall
and weaker operation contracts cause the largest losses within
the typed pipeline.

Free-Text Stages preserves the four-stage order,
stage-specific prompts, closed operation documentation,
agent backbone, decoding configuration, and evaluation
protocol. It replaces the typed executable state and
deterministic compilation boundary with prose intermediates
followed by direct CadQuery realization. This comparison
therefore evaluates whether staged decomposition remains
sufficient when intermediate decisions are not maintained as
typed executable commitments. It should be interpreted as a
boundary-level comparison instead of an isolated ablation of
any single schema or compiler component.

Table~\ref{tab:app_free_text_alignment} records the matched
and removed components of this comparison.

\begin{center}
\begin{minipage}{\columnwidth}
\centering
\small
\setlength{\tabcolsep}{2pt}%
\begin{tabularx}{\columnwidth}{@{}p{0.22\columnwidth}p{0.40\columnwidth}Y@{}}
\toprule
Aspect & Matched configuration & Removed \\
\midrule
Skeleton and prompts
& Same four-stage skeleton and stage-specific prompt source
& Schema constraints only \\
Operation documentation
& Same closed operation documentation
& -- \\
Backbone and decoding
& Qwen3.5-9B, temperature 0.0, and top-p 1.0
& -- \\
Evaluation
& Same frozen ground truth, gates, checkers, and denominator rules
& -- \\
Execution boundary
& Same four-stage order and CadQuery target
& Typed ledger, node schema, typed relations, parameter graph,
and compiler \\
\bottomrule
\end{tabularx}
\captionof{table}{Alignment of Free-Text Stages with the full pipeline. The
removed elements all belong to the typed state and deterministic
compilation boundary.}
\label{tab:app_free_text_alignment}
\end{minipage}
\end{center}

Table~\ref{tab:app_free_text_failures} reports the observed
outcome counts for the same 300 prompts.

\begin{center}
\begin{minipage}{\columnwidth}
\centering
\small
\setlength{\tabcolsep}{3.5pt}
\begin{tabularx}{\columnwidth}{@{}Yr@{}}
\toprule
Outcome & Cases \\
\midrule
Code generation failure & 39 \\
Execution failure & 251 \\
Success & 10 \\
\bottomrule
\end{tabularx}
\captionof{table}{Failure attribution for Free-Text Stages over 300 MHCAD
prompts.}
\label{tab:app_free_text_failures}
\end{minipage}
\end{center}

All four stages return outputs for 300/300 prompts, but only
10 cases produce valid STEP files. The resulting 0.9\% SY
shows that, under the matched staged workflow, prose
intermediates followed by direct code realization do not
provide the delivery reliability of the typed executable
pipeline. Among the remaining outcomes, 39 fail during code
generation and 251 fail during execution.

\subsection{Results Across Agent Backbones}

Only the agent backbone is changed in this study. The
operation library, prompts, compiler, CAD kernel, and
structural evaluation remain fixed.

\begin{center}
\begin{minipage}{\columnwidth}
\centering
\small
\setlength{\tabcolsep}{1.6pt}%
\begin{tabularx}{\columnwidth}{@{}Yccc@{}}
\toprule
Agent backbone
& \shortstack{STEP/B-rep\\valid}
& \shortstack{SF\\95\% CI}
& \shortstack{SY\\95\% CI} \\
\midrule
Qwen3.5-9B
& \shortstack{270/300\\(90.0\%)}
& \shortstack{89.8\\[0.2ex] [86.2, 93.1]}
& \shortstack{82.7\\[0.2ex] [78.0, 87.1]} \\

Qwen3.6-27B
& \shortstack{295/300\\(98.3\%)}
& \shortstack{87.0\\[0.2ex] [82.9, 90.7]}
& \shortstack{86.0\\[0.2ex] [81.7, 89.9]} \\

Phi-4-14B
& \shortstack{295/300\\(98.3\%)}
& \shortstack{89.3\\[0.2ex] [85.5, 92.6]}
& \shortstack{87.4\\[0.2ex] [83.3, 91.1]} \\

Granite 4.1 8B
& \shortstack{294/300\\(98.0\%)}
& \shortstack{86.5\\[0.2ex] [82.6, 90.2]}
& \shortstack{84.5\\[0.2ex] [80.1, 88.6]} \\

Nemotron-Nano-12B-v2
& \shortstack{253/300\\(84.3\%)}
& \shortstack{93.6\\[0.2ex] [90.4, 96.5]}
& \shortstack{77.2\\[0.2ex] [70.8, 83.3]} \\
\bottomrule
\end{tabularx}
\captionof{table}{Results across agent backbones with bootstrap confidence
intervals. STEP and B-rep-valid counts coincide for every backbone;
the final row is NVIDIA Nemotron-Nano-12B-v2.}
\label{tab:app_agent_backbones}
\end{minipage}
\end{center}

All five backbones use the same structured interface.
Differences between SF and SY reflect valid-output coverage.


\section{Generalization to CADPrompt}
\label{app:cadprompt}

CADPrompt is used as an independently authored prompt source.
The main paper reports the compact structural comparison; this
section documents subset construction and complete counts.

\subsection{Subset Construction}

The structural subset of 30 cases is selected using prompt
scope and checker applicability, independently of model outputs
and evaluation results. Table~\ref{tab:app_cadprompt_filtering}
summarizes the retained cases and applicable checks.

\begin{appcoltable}
\begin{tabularx}{\columnwidth}{@{}p{0.27\columnwidth}Yr@{}}
\toprule
Group & Item & Count \\
\midrule
Overall & Total CADPrompt cases & 200 \\
Overall & Kept single mechanical structural cases & 30 \\
Overall & Removed & 170 \\
Kept family & Bored cylindrical part & 8 \\
Kept family & Fastener like single part & 1 \\
Kept family & Pipe or flange like single part & 5 \\
Kept family & Plate with holes & 3 \\
Kept family & Ring, washer, or bushing like & 13 \\
Removed reason & No applicable validated structural check & 165 \\
Removed reason & Outside single standard part scope & 3 \\
Removed reason & Abstract sketch profile & 1 \\
Removed reason & Nonmechanical furniture or toy like object & 1 \\
Validated check & Central bore presence & 23 \\
Validated check & Coaxial annular axis & 18 \\
Validated check & Mounting hole count & 3 \\
Validated check & Simple circular pattern & 3 \\
\bottomrule
\end{tabularx}
\caption{Construction of the 30 case CADPrompt structural subset.}
\label{tab:app_cadprompt_filtering}
\end{appcoltable}

\subsection{Exact Results}

Table~\ref{tab:app_cadprompt_counts} gives the complete
artifact and structural counts for the selected subset.

\begin{appcoltable}
\setlength{\tabcolsep}{1.8pt}%
\begin{tabularx}{\columnwidth}{@{}Ycccc@{}}
\toprule
Method & \shortstack{STEP found/\\reimp./cases} & \shortstack{B-rep\\valid} & SF & SY \\
\midrule
\method{} (Ours) & 30/30/30 & 30/30 & 41/47 & 41/47 \\
Text2CAD & 30/30/30 & 30/30 & 17/47 & 17/47 \\
Text-to-CQ & 5/5/30 & 5/30 & 3/8 & 3/47 \\
CADFusion & 26/26/30 & 26/30 & 32/40 & 32/47 \\
Qwen Direct & 25/25/30 & 25/30 & 24/41 & 24/47 \\
\bottomrule
\end{tabularx}
\caption{Exact CADPrompt results on the structural subset of
30 cases. SF and SY percentages appear in the main-paper subset
figure.}
\label{tab:app_cadprompt_counts}
\end{appcoltable}

\subsection{Results by Structural Requirement}

Table~\ref{tab:app_cadprompt_percheck} provides the available
per-check breakdowns behind the aggregate comparison.

\begin{appcoltable}
\setlength{\tabcolsep}{1.8pt}%
\begin{tabularx}{\columnwidth}{@{}Ycccc@{}}
\toprule
Method & Bore & \shortstack{Coaxial\\Axis} & \shortstack{Mount\\Holes} & \shortstack{Circular\\Pattern} \\
\midrule
\method{} (Ours) & 22/23 & 17/18 & 1/3 & 1/3 \\
Text2CAD & 10/23 & 7/18 & 0/3 & 0/3 \\
Text-to-CQ & 2/3 & 1/3 & 0/1 & 0/1 \\
CADFusion & 17/19 & 11/15 & 2/3 & 2/3 \\
\bottomrule
\end{tabularx}
\caption{Available CADPrompt passed/eligible counts by structural
requirement. Aggregate Qwen Direct results are reported in
Table~\ref{tab:app_cadprompt_counts}.}
\label{tab:app_cadprompt_percheck}
\end{appcoltable}

The transferred subset contains 47 checks. \method{} passes
41, including 22 of 23 bore checks and 17 of 18 coaxial axis
checks. The mounting hole and pattern groups each contain
only three checks and should be interpreted with their small
denominators.


\FloatBarrier
\section{Implementation Contracts and Algorithms}
\label{app:implementation}

This section specifies the structured interfaces and
deterministic procedures used by \method{}. Design, Recall, and
Relation use structured agent outputs. Parameter completion and
CAD realization follow procedures defined for each operation.

\subsection{Stage Contracts}

\begin{center}
\begin{minipage}{\columnwidth}
\centering
\small
\setlength{\tabcolsep}{3.5pt}
\begin{tabularx}{\columnwidth}{@{}p{0.18\columnwidth}Y@{}}
\toprule
Stage & Interface contract \\
\midrule
Design
& \textit{Input:} requirement and benchmark metadata.
\textit{Output:} construction intent record $D$.
Preserve dimensions, counts, visible roles, and relation cues;
leave operation selection to Recall. \\

Recall
& \textit{Input:} $D$ and allowed operations.
\textit{Output:} operation nodes $G_O$.
Select operations from the library and link every node to
requirement evidence. \\

Relation
& \textit{Input:} $D$, operation nodes, and relation cues.
\textit{Output:} typed relation graph $G_R$.
Use supported relation types and make ownership, alignment,
containment, and repeated layout explicit. \\

Parameters
& \textit{Input:} $D$, $G_R$, operation interfaces, and value
domains. \textit{Output:} parameterized graph $G_P$.
Resolve required values with deterministic operation rules and
retain unresolved values as typed errors. \\
\bottomrule
\end{tabularx}
\captionof{table}{Contracts for the four GOPR-CAD stages.}
\label{tab:app_stage_contracts}
\end{minipage}
\end{center}

Table~\ref{tab:app_stage_contracts} separates each stage's
input, structured output, and validity conditions. The
implementation labels are \texttt{on\_face},
\texttt{coaxial}, \texttt{patterned\_about},
\texttt{contains}, and \texttt{adjacent}. In the pipeline
figure, \texttt{hosted\_on}, \texttt{coaxial\_with}, and
\texttt{share\_layout} are readable aliases for host-face,
coaxial, and repeated-layout relations. Relation edges resolve
hosts, selectors, execution dependencies, and parameter
propagation before compilation. The compiler then consumes
the resolved operation identities, selector bindings, and
parameter dictionaries.

\subsection{Structured Output Schemas}

\begin{center}
\begin{minipage}{\columnwidth}
\centering
\small
\setlength{\tabcolsep}{3.5pt}
\begin{tabularx}{\columnwidth}{@{}p{0.20\columnwidth}Y@{}}
\toprule
Stage & Compact schema \\
\midrule
Design
& \texttt{\{}\,\texttt{part\_summary}, \texttt{visible\_mechanical\_roles},
\texttt{construction\_outline}, \texttt{relation\_hints},
\texttt{parameter\_hints}, \texttt{uncertain\_items}\,\texttt{\}} \\

Recall
& \texttt{\{}\,\texttt{operation\_nodes}, \texttt{typed\_missing\_roles},
\texttt{coverage\_audit}\,\texttt{\}} \\

Relation
& \texttt{\{}\,\texttt{relations}, \texttt{missing\_required\_relations},
\texttt{relation\_audit}\,\texttt{\}} \\

Parameters
& \texttt{\{}\,\texttt{parameter\_bindings},
\texttt{typed\_parameter\_errors}\,\texttt{\}} \\
\bottomrule
\end{tabularx}
\captionof{table}{Compact schemas for the four GOPR-CAD stages.}
\label{tab:app_stage_schemas}
\end{minipage}
\end{center}

\subsection{GOPR-CAD Inference Algorithm}

\begin{algorithm}[H]
\caption{GOPR-CAD inference for one requirement}
\label{alg:app_gopr_inference}
\small
\begin{algorithmic}[1]
\REQUIRE Requirement $x$, operation library $\mathcal{O}$,
compiler $\mathcal{C}$, CAD kernel $\mathcal{K}$
\ENSURE CAD model $y$ or a structured failure record
\STATE $D \gets \Phi_D^{\theta}(x)$
\COMMENT{represent construction intent}
\STATE $G_O \gets \Phi_O^{\theta}(D,\mathcal{O})$
\COMMENT{select executable operations}
\IF{a required construction role has no supported operation}
    \STATE record a typed missing role
\ENDIF
\STATE $G_R \gets \Phi_R^{\theta}(D,G_O)$
\COMMENT{add supported mechanical relations}
\STATE $P \gets \emptyset$
\FOR{dependency level $\ell=0$ to $L$}
    \STATE $P^{(\ell)} \gets
    \Phi_P^{(\ell)}(D,G_R,V^{(\ell)},P^{(<\ell)})$
    \STATE $P \gets P\cup P^{(\ell)}$
\ENDFOR
\STATE $G_P \gets (V_O,E_R,P,\alpha)$
\IF{$G_P$ does not satisfy operation contracts or parameter completeness}
    \STATE \textbf{return} structured graph failure record
\ENDIF
\STATE $\Pi \gets \mathcal{C}(G_P)$
\STATE $y \gets \mathcal{K}(\Pi)$
\IF{execution succeeds}
    \STATE export CAD artifacts
    \STATE \textbf{return} $y$
\ELSE
    \STATE \textbf{return} structured execution failure record
\ENDIF
\end{algorithmic}
\end{algorithm}

Given the completed graph state, CAD realization is fully
deterministic.

\subsection{Structural Evaluation Algorithm}

\begin{algorithm}[H]
\caption{STEP/B-rep structural evaluation}
\label{alg:app_structural_eval}
\small
\begin{algorithmic}[1]
\REQUIRE Generated artifacts, structural records
$\{\mathcal{A}_i\}$, CAD kernel, structural checkers
\ENSURE STEP, SF, SY, and diagnostic counts
\FOR{each prompt $i$}
    \STATE determine whether a STEP file is present
    \IF{STEP is missing}
        \STATE set $v_i\gets 0$ and all applicable passes to zero
        \STATE \textbf{continue}
    \ENDIF
    \STATE reimport the STEP file and test B-rep validity
    \IF{the B-rep is invalid}
        \STATE set $v_i\gets 0$ and all applicable passes to zero
        \STATE \textbf{continue}
    \ENDIF
    \STATE set $v_i\gets 1$
    \FOR{each $j\in\mathcal{A}_i$}
        \STATE run checker $j$ with the shared tolerance settings
        \STATE store $\mathrm{pass}_{ij}\in\{0,1\}$
    \ENDFOR
\ENDFOR
\STATE compute SF over checks attached to valid B-reps
\STATE compute SY over all applicable checks
\STATE report per-check numerators, denominators, and output failures
\end{algorithmic}
\end{algorithm}

The evaluator computes SF and SY exactly as defined in
Section~\ref{app:structural_eval} while retaining the
diagnostic counts reported in Appendix~\ref{app:mhcad_results}.

\end{document}